\documentclass[
  manuscript=article,  
  layout=preprint,  
  year=2023
]{extra/joas}

\conference{Conference Title}

\received {}
\revised  {}
\accepted {}
\published{}

\editor{Editor Name}

\reviewers{First Reviewer, Second Reviewer, Third Reviewer}

\usepackage[english]{babel} 
\usepackage{blindtext}

\title{Multimodal Short Video Rumor Detection System Based on Contrastive Learning}

\author{Yuxing Yang}
\affiliation{Sichuan University Cyberspace Security Academy, Chengdu, China}

\author{Junhao Zhao}
\affiliation{Sichuan University Cyberspace Security Academy, Chengdu, China}

\author{Siyi Wang}
\affiliation{Sichuan University Cyberspace Security Academy, Chengdu, China}

\author{Xiangyu Min}
\affiliation{Sichuan University Cyberspace Security Academy, Chengdu, China}

\author{Pengchao Wang}
\affiliation{Sichuan University Cyberspace Security Academy, Chengdu, China}

\author{Haizhou Wang}
\affiliation{Sichuan University Cyberspace Security Academy, Chengdu, China}
\email{whzh.nc@scu.edu.cn}


\begin{document}

\begin{abstract}
With the rise of short video platforms as prominent channels for news dissemination, major platforms in China have gradually evolved into fertile grounds for the proliferation of fake news. However, distinguishing short video rumors poses a significant challenge due to the substantial amount of information and shared features among videos, resulting in homogeneity. To address the dissemination of short video rumors effectively, our research group proposes a methodology encompassing multimodal feature fusion and the integration of external knowledge, considering the merits and drawbacks of each algorithm. The proposed detection approach entails the following steps: (1) creation of a comprehensive dataset comprising multiple features extracted from short videos; (2) development of a multimodal rumor detection model: first, we employ the Temporal Segment Networks (TSN) video coding model to extract video features, followed by the utilization of Optical Character Recognition (OCR) and Automatic Speech Recognition (ASR) to extract textual features. Subsequently, the BERT model is employed to fuse textual and video features; (3) distinction is achieved through contrast learning: we acquire external knowledge by crawling relevant sources and leverage a vector database to incorporate this knowledge into the classification output. Our research process is driven by practical considerations, and the knowledge derived from this study will hold significant value in practical scenarios, such as short video rumor identification and the management of social opinions.
\end{abstract}

\section{Introduction}
With the popularity of short video platforms such as Douyin and Kuaishou, they have become an important channel for news sharing. Unlike traditional news dissemination dominated by professional news media, on short video platforms, not only can official news media publish short videos, but ordinary people can also post various types of short videos. However, this openness has also become an important prerequisite for short video platforms to evolve into a breeding ground for fake news. In addition, due to the many characteristics of short videos, they have stronger dissemination power than traditional rumors. In terms of dissemination methods, short videos are disseminated visually, which can bring stronger visual impact and emotional shock to the public. In terms of the use of dissemination techniques, short video rumors mainly use methods that appeal to emotions and fear, create a certain atmosphere, arouse the audience's crisis awareness and nervousness, and use the audience's psychology of maintaining social order, maintaining mainstream values, and touching the general public's universal interests to keep the audience continuously interested in the content of the rumor. In terms of content dissemination, short video rumors mainly revolve around public social events and livelihood issues that concern the general public. Livelihood issues have always been the most concerned issue among the public, and they are easily prone to causing public opinion. The words "Internet celebrity" and the like have aroused widespread social attention. Therefore, studying fake news in video format is of great significance for timely discovering and monitoring the spread of fake news.

In the task of detecting rumors in short videos, the most challenging part is how to extract features from the content to identify whether a post is a rumor. Currently, there are roughly two types of rumor detection methods: Propagation-Based Rumor Detection and Content-Based Rumor Detection. Propagation-based methods extract features for post classification based on user responses received on social media. It can further be divided into propagation pattern analysis and response user analysis. Content-based methods extract features from the content of the post being verified for post classification. Depending on the modality of the post content, it can be further divided into single-modality and multi-modality rumor detection methods.

Single-modality rumor detection methods, such as traditional machine learning methods like Support Vector Machine (SVM), decision trees, and random forests, rely heavily on manual feature extraction and are time-consuming. For example, SVM-TS uses heuristic rules and linear SVM to classify rumors on Twitter, and uses a time-series structure to simulate changes in social features. Multi-modality rumor detection methods have gained popularity in recent years due to the development of multimedia technology. Post content has evolved from pure text to multi-modal content including text, images, and videos. With the exceptional performance of deep neural networks (DNN) in non-linear representation learning, many multi-modal representation methods use deep learning to learn feature representations of post content and have achieved superior performance in rumor detection. A new attention mechanism recursive neural network (att-RNN) has been proposed to fuse multi-modal features for effective rumor detection.

In addition to the aforementioned research methods, some researchers have utilized pre-trained models such as Bidirectional Encoder Representation from Transformers (BERT)\cite{devlin2018bert} to extract better text multi-level representations for performing rumor detection tasks. For example, some works have explored the architecture of the BERT model, extracting multi-level semantic representations from the intermediate hidden layers of BERT, which has been shown to improve the performance of the model for better execution of downstream tasks.

Although existing work has achieved some achievements in rumor video detection, they only utilize the available modalities in the experimental dataset for detection, inevitably missing some important clues. Moreover, because these methods are evaluated on different datasets, they lack effective comparisons between them. Therefore, in this work, we will establish a multi-feature video dataset and propose a new multi-modal detection model.

Based on previous research, this project proposes a new multimodal short-video rumor detection system using contrastive learning. The general framework of the project is as follows: 1) Establishment of dataset: Building a rumor short-video dataset with multiple features; 2) Multimodal rumor detection model: Firstly, using the TSN (Temporal Segment Networks) video encoding model to extract video visual features; then using the fusion of Optical Character Recognition (OCR) and Automatic Speech Recognition (ASR) to extract text; then using the BERT model to fuse text features and video visual features; finally, using contrastive learning to differentiate: Firstly, crawling external knowledge, then using vector databases to introduce external knowledge and output the final classification structure.

\section{Overall model architecture}

\begin{figure}[H]
\centering
\includegraphics[scale=.2]{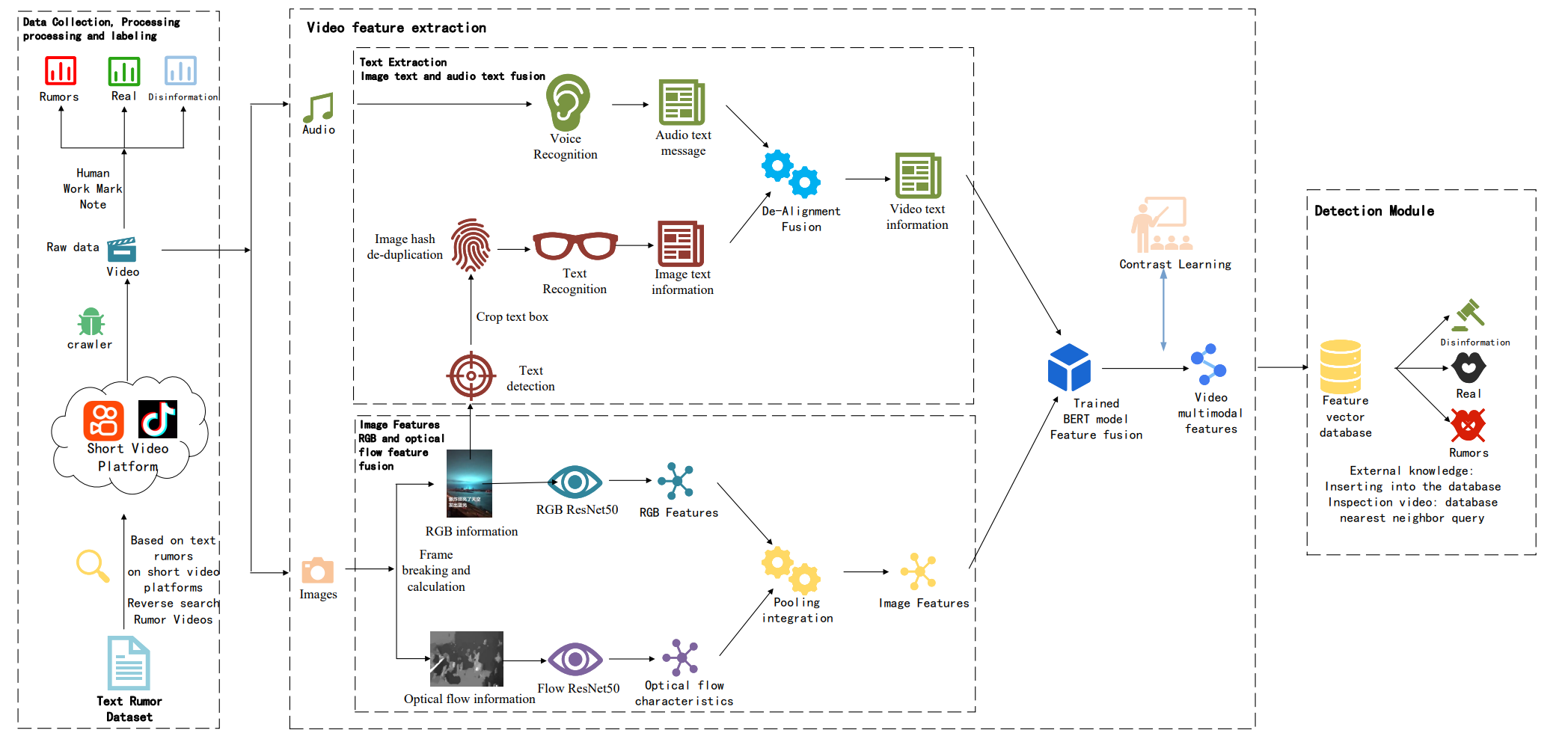}\\
\caption{Overall model architecture}
\end{figure}

\subsection{Data collection, processing and labeling}

The first part of this project is data collection, processing, and annotation, which mainly includes the following three steps:
\begin{enumerate}\renewcommand{\labelenumi}{(\theenumi).}
\item Using Python web crawlers with common text rumor datasets as keywords, search for short videos in the field where rumors may appear, such as knowledge or news, on common short video platforms such as Douyin by reverse searching for rumor keywords, and download them.

\item Process the collected rumor dataset and manually label the rumor data.

\item Use crawlers to scrape official or credible knowledge platforms such as China Refutes Rumors and Baidu Encyclopedia as truth, divide the original articles into knowledge paragraphs, extract their text features to obtain their feature vectors, and insert them into the database.
\end{enumerate}
\subsection{Feature extraction}

The second part of this project is the multimodal feature extraction module, which mainly includes the following five steps:
\begin{enumerate}\renewcommand{\labelenumi}{(\theenumi).}
\item Video image feature processing. Frames are extracted from the video, and the ResNet50 neural network model is used to process the raw information of RGB and optical flow extracted from the video through the RGB model and optical flow model, respectively, to obtain RGB and optical flow features. The two features are pooled and fused as the image features of the video.

\item Subtitle text detection. For the subtitles in the short video, text boxes can be extracted by positioning and cropping, and similar frames with high similarity are filtered using efficient algorithms such as perceptual hashing for adjacent content with the same subtitles. Finally, the optical character recognition (OCR) is used to convert the subtitle images into text.

\item Audio processing. For the audio part extracted from the short video, VSD speech endpoint detection is used to detect the start and end timestamps of each segment of effective speech, and only the detected segments of effective speech are retained as the input for subsequent recognition engines to filter out invalid sound fragments and reduce recognition errors in subsequent steps. The core step of audio processing is to use automatic speech recognition (ASR) technology to convert each segment of audio speech into text. To increase the readability of the text recognized by the speech recognition module during manual debugging and retain some information such as sentence breaks and emotions for downstream steps, a punctuation prediction step is performed after speech recognition.

\item Fusion of image text and audio text. Based on the recognized information such as the start and end timestamps of each sentence and the image text information extracted from the subtitles, they are merged and aligned. Similarities are judged by the edit distance for deduplication, and the video text information is obtained.

\item Fusion of video text information and image information. The encoded video text information and image information are input into the Bert model, and the cross-attention mechanism of the Bert model is used for fusion.
\end{enumerate}

\subsection{Comparing learning and label prediction}

The third part of this project is contrastive learning to train the Bert model and predict labels.

As judging rumors is difficult without external support, we consider introducing external knowledge and storing the extracted feature of known knowledge projects in a vector database. The final prediction result is obtained by retrieving the feature of the query video in the vector database. This makes it easy to add knowledge in real-time and ensures the accuracy of the model.

In order to achieve excellent vector search performance, we need to calculate the contrastive loss of the model for contrastive learning, pulling the samples with the same labels closer and pushing those with different labels apart, and achieving data clustering and differentiation.

The feature extraction model is used to extract feature vectors from the query video. The nearest 10 vectors in the feature vector database are retrieved through the nearest neighbor query. The label is accumulated with the similarity as the weight, and the highest weighted score is used as the predicted value for the video type judgment.

\section{Model specific implementation}

\subsection{Image Feature Extraction}

In 2014, Simonyan et al.\cite{simonyan2014two} proposed a dual-stream network model using convolutional neural networks (CNNs), which fused the classification results of two branches: spatial stream and temporal stream, to effectively extract spatial and temporal features. Through a series of convolutional and fully connected layers, the RGB image and optical flow were combined, and a softmax classifier was used to output the prediction result. Finally, the scores of the two network streams were fused to obtain the final classification performance.

Due to the limited access to the contextual temporal relationship, traditional two-stream convolutional networks can only process single data in a single frame or video segment. Therefore, Wang et al.\cite{wang2016temporal} proposed a network called TSN that combines video-level supervision with a sparse sampling strategy for video frames. TSN is based on a two-stream network, but with a different approach - it uses a sparse sampling method and makes an initial prediction of the action type for each video segment. Then, by using a segmental consensus function, it obtains a video-level prediction. Finally, all the predictions are combined to produce the final prediction result.

Due to the simplicity and efficiency of TSN, this paper uses TSN to extract image features.

First, the video is broken down into frames, obtaining \emph{RGB frames} and \emph{optical flow frames}. The \emph{optical flow} is used to describe the objects in a scene that undergo dynamic changes between two consecutive frames due to motion (either of the object or of the camera). Its essence is a \emph{two-dimensional vector field}, where each vector represents the displacement of the point in the scene from the previous frame to the current frame.

\begin{figure}[H]
\centering
\includegraphics[scale=.4]{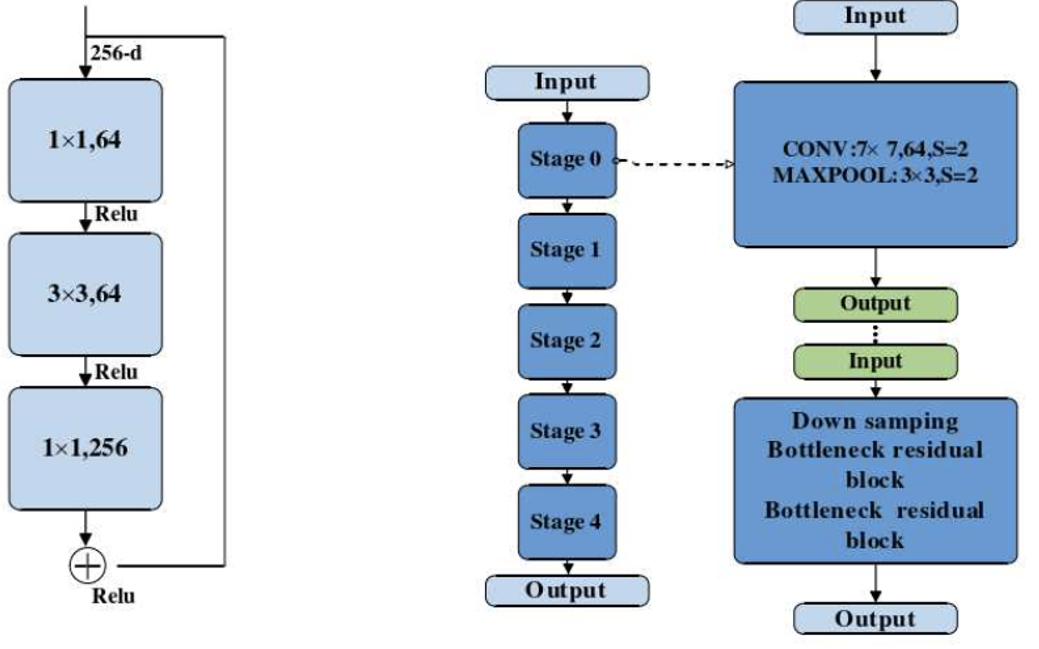}\\
\caption{Results of the bottleneck residual module and the network structure of ResNet-50}\label{ResNet}
\end{figure}

Then, the RGB frames and the optical flow frames are separately fed through pre-trained \emph{ResNet-50} networks to obtain RGB features and optical flow features. This paper uses ResNet-50 as the basic feature extraction network for motion information features. The most typical improvement of ResNet-50 is the design of bottleneck residual blocks. As shown in the figure \ref{ResNet}, a 3×3×256 feature vector is input, then 64 1x1 convolutions are used to reduce the 256 channels to 64, and finally 1x1 convolution is used to restore 256 channels. Stage 0~Stage4 represent $conv1, conv2_X,conv3_x, conv4_x, conv5_x$, respectively, and each layer performs downsampling with step s of 2, where $Convl_x$ mainly performs convolution operation, normalization operation, activation function calculation, and max pooling operation on input values. $conv2_x. Conv3_x, Conv4_x$ and $Conv4_x$ all contain residual blocks, each of which has three convolution layers, i.e., 18-layer, 34-layer, and 50-layer. After a series of convolutional operations, motion features can be successfully extracted. Finally, before outputting the results, average pooling is performed to convert the extracted motion features into feature vectors. The softmax classifier performs operations on each feature vector and outputs the probability of action categories. Therefore, to balance performance and computation cost, this paper uses the ResNet-50 network model as one of the basic feature extraction network models. Through the ResNet50 network, we obtain two feature vectors, each with 200 dimensions.

The two features are then concatenated to form a 400-dimensional feature vector, and all frames are transformed into a $100 \times 400$ video feature through an \emph{average pooling layer}.

\subsection{Text feature extraction}

\subsubsection{Speech recognition}

We use the efficient non-autoregressive end-to-end speech recognition framework, Paraformer\cite{gao2022paraformer}. The Paraformer model consists of five components: Encoder, Predictor, Sampler, Decoder, and Loss function, as shown in the figure below. The Encoder can adopt different network structures such as self-attention, conformer, and SAN-M. The Predictor has two layers of FFN, predicting the number of target words and extracting the corresponding acoustic vectors. The Sampler is a module without learnable parameters that produces semantically meaningful feature vectors based on the input acoustic vectors and target vectors. The Decoder structure is similar to autoregressive models, except that it is bidirectional (autoregressive is unidirectional). In addition to the cross-entropy (CE) and MWER discriminative optimization targets, the Loss function section also includes the Predictor optimization target MAE.

\textbf{Speech endpoint detection}

FSMN-Monophone VAD is a speech endpoint detection model used to detect the starting and ending time points of valid speech in input audio, and input the detected valid audio segments into the recognition engine for recognition, reducing recognition errors caused by invalid speech.

\begin{figure}[H]
\centering
\includegraphics[scale=.6]{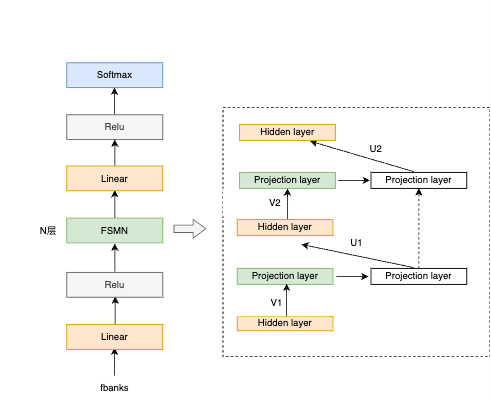}\\
\caption{FSMN-Monophone VAD model structure}\label{VAD}
\end{figure}

The structure of the FSMN-Monophone VAD model is shown in the figure \ref{VAD}: at the model structure level, the FSMN model structure can consider contextual information, has fast training and inference speed, and the latency is controllable; at the same time, according to the requirements of VAD model size and low latency, the network structure and the number of right context frames of the FSMN have been adapted. At the modeling unit level, speech information is rich, and the learning ability of using a single class to represent it is limited, so a single speech class is upgraded to Monophone. The subdivision of modeling units can avoid parameter averaging, enhance abstract learning ability, and improve discriminability.

\textbf{Automatic speech recognition}

We use the efficient non-autoregressive end-to-end speech recognition framework, Paraformer. The Paraformer model consists of five components: Encoder, Predictor, Sampler, Decoder, and Loss function, as shown in the figure below. The Encoder can adopt different network structures such as self-attention, conformer, and SAN-M. The Predictor has two layers of FFN, predicting the number of target words and extracting the corresponding acoustic vectors. The Sampler is a module without learnable parameters that produces semantically meaningful feature vectors based on the input acoustic vectors and target vectors. The Decoder structure is similar to autoregressive models, except that it is bidirectional (autoregressive is unidirectional). In addition to the cross-entropy (CE) and MWER discriminative optimization targets, the Loss function section also includes the Predictor optimization target MAE.

\begin{figure}[H]
\centering
\includegraphics[scale=.6]{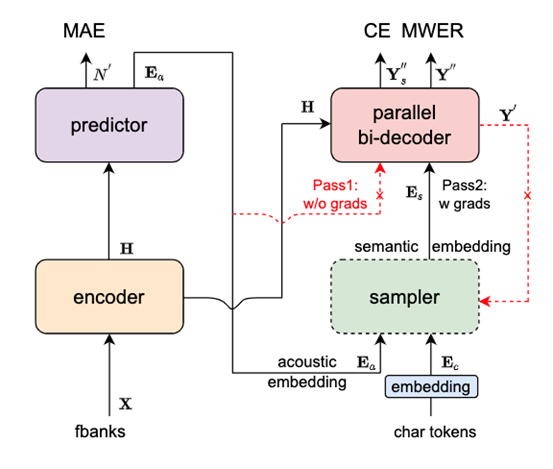}\\
\caption{Paraformer model structure}
\end{figure}

The main points are:

The Predictor module: Based on Continuous integrate-and-fire (CIF), it extracts acoustic feature vectors corresponding to target characters, which can more accurately predict the number of target characters in the speech.

The Sampler: By sampling, acoustic feature vectors and target word vectors are transformed into feature vectors containing semantic information, which are combined with a bidirectional decoder to enhance the model's ability to model context. The MWER training criterion is based on negative sampling.

\textbf{Punctuation prediction}

We use the punctuation module in the Controllable Time-delay Transformer post-processing framework. It predicts punctuation for the recognized speech-to-text results, making the text more readable.

\begin{figure}[H]
\centering
\includegraphics[scale=.8]{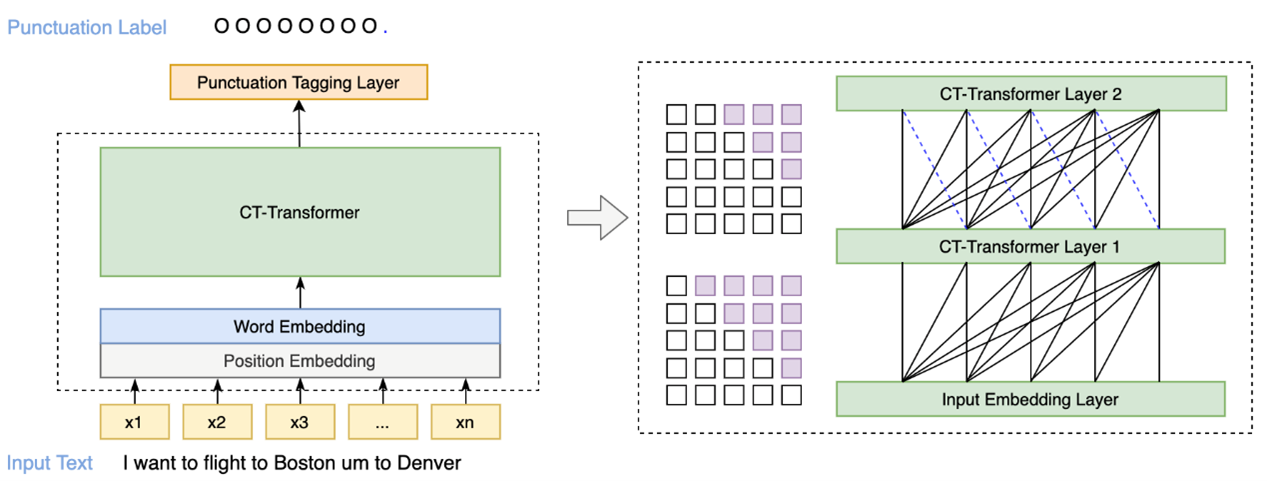}\\
\caption{Controllable Time-delay Transformer model structure}\label{CTT}
\end{figure}

The structure of the Controllable Time-delay Transformer model is shown in the figure \ref{CTT}, which consists of three parts: Embedding, Encoder, and Predictor. Embedding is a word vector with a positional vector added. Encoder can use different network structures, such as self-attention, conformer, SAN-M, etc. Predictor predicts the punctuation type after each token.

We chose the high-performance Transformer model for our system. While the Transformer model achieves good performance, its self-serializing input characteristic leads to significant latency in the system. The conventional Transformer model can see all future information, which causes punctuation to depend on far future information. This creates an unpleasant feeling for users, where punctuation keeps changing and refreshing, resulting in unstable results over a long time. To address this issue, we proposed an innovative Controllable Time-Delay Transformer (CT-Transformer) model that effectively controls punctuation delay without sacrificing model performance.

\subsubsection{Image text recognition}

Firstly, \emph{perceptual hash (phash)} \cite{niu2008overview} is calculated for each video frame to obtain the image similarity, and similar frames are filtered out. Then, text detection is performed on the filtered frames, and perceptual hash (phash) calculation is performed again for each text box image to filter out similar text boxes. Finally, text recognition is performed to obtain the text in the image.

The implementation steps of \emph{perceptual hash (phash)} are as follows:
\begin{enumerate}\renewcommand{\labelenumi}{(\theenumi).}
\item The image is downscaled to remove high-frequency and detailed information, while retaining the structural brightness and darkness of the image. The image is reduced to an 8x8 size, which is a total of 64 pixels, eliminating the differences in images caused by different sizes and proportions.
\item The color is simplified. The downscaled image is converted to 64-level grayscale. In other words, there are only 64 colors for all pixels.
\item The Discrete Cosine Transform (DCT) is calculated. The DCT divides the image into frequency clusters and stepped shapes. Although JPEG uses an 8x8 DCT transformation, a 32x32 DCT transformation is used here.

\item The DCT is further downscaled. Although the result of the DCT is a 32x32 matrix, only the top-left 8x8 matrix is retained, which presents the lowest frequency of the image.

\item The average value is calculated. The average value of all 64 values is calculated.

\item The DCT is further reduced to calculate the hash value. This is the most important step. Based on the 8x8 DCT matrix, a 64-bit hash value is set to "1" if it is greater than or equal to the mean value of the DCT, and "0" if it is less than the mean value of the DCT.
\end{enumerate}

\begin{figure}[H]
\centering
\includegraphics[scale=.2]{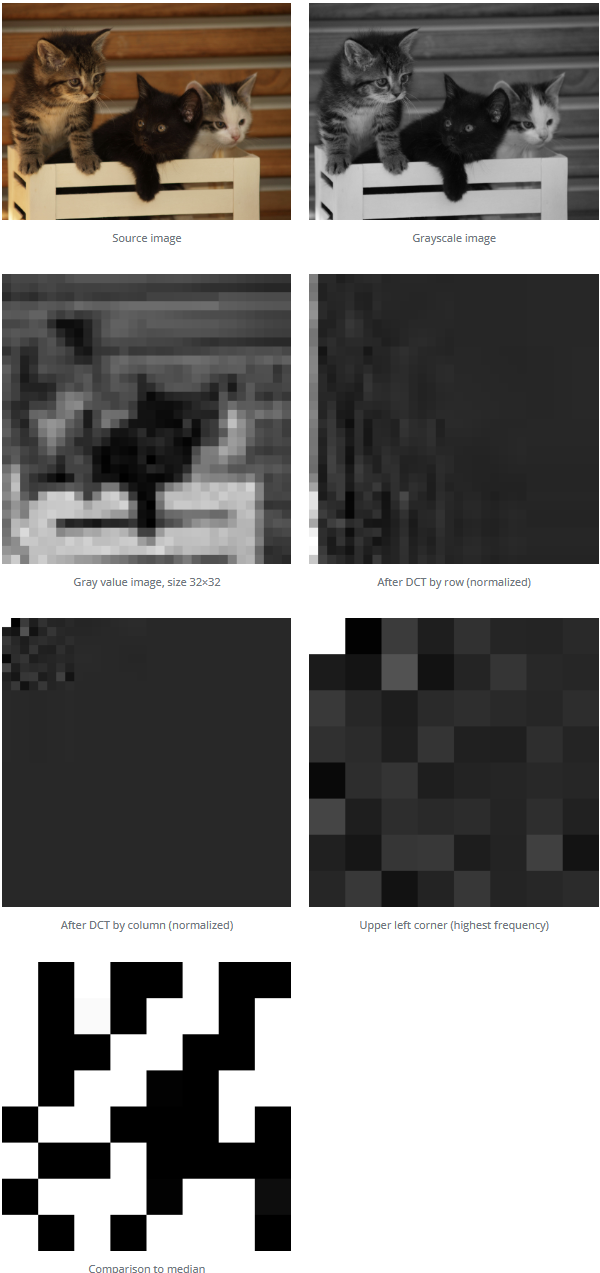}\\
\caption{The effect of each step of phash}
\end{figure}

Specifically, for each image, after phash calculation, we obtain a 64-bit binary hash value. If the number of different bits between two hash values is less than 8, we consider these two images as similar; otherwise, they are not similar. For each frame of the video, we compare it with the adjacent 10 frames. If all of them are similar, we discard the redundant frames; otherwise, we keep them. After that, we perform text detection on the remaining frames to obtain several text box images. We conduct a similar deduplication process for these text box images by comparing them with the adjacent 100 text box images while preserving their timestamps. Finally, we perform text recognition on the text box images to obtain the image text.

\subsubsection{Audio text and image text fusion}

Through merging and alignment based on the timestamps corresponding to the audio text and image text, we use two pointers to traverse the audio text and image text, and if the timestamp of one pointer is smaller than the timestamp of the other pointer, we process the data corresponding to that pointer. For each statement to be added, we compare its similarity to the added statements using \emph{edit distance}, also called Levenshtein distance. We remove the image text with high similarity to achieve text deduplication. The results are then concatenated to obtain the video text.

Calculation of \emph{edit distance}: The Levenshtein distance is the minimum number of edit operations required to transform one string into another. We use dynamic programming algorithm to calculate it.

\subsection{Image and text information integration}

This article is based on a fusion mechanism that prioritizes text modality and supplements with video modality. The article uses BERT (bidirectional encoder representation from transformers) model to combine a large amount of text and video image information, in order to achieve the effect of text-based classification.

BERT, short for Bidirectional Encoder Representation from Transformers, is a pre-trained language representation model. It emphasizes no longer using traditional unidirectional language models or shallow concatenation of two unidirectional language models for pre-training, but instead adopting a new approach called masked language model (MLM) to generate deep bidirectional language representations.

As the input feature dimension of the BERT model is (512, 768), which requires 512 vectors of 768 dimensions, we allocate one vector for cls, $m$ vectors for video features, and $520-m$ vectors for text features. According to subsequent experiments, the best result is achieved when $m$ is 25. It can be seen that the text modality feature plays a dominant role in the fusion process, while the video modality feature plays a complementary and corrective role in the fusion.

The BERT input mainly consists of video information and textual information. For video information, the output of TSN (video feature extraction model) with size of $100 \times 400$ is averaged-pooled to obtain a $25 \times 768$ vector. For textual information, BERT model is used to convert text information into a $485\times 768$ word vector. The above two aspects constitute the input of the BERT model, plus a \emph{CLS} and \emph{SEP} token to form all the inputs of this project.

\begin{figure}[H]
\centering
\includegraphics[scale=.3]{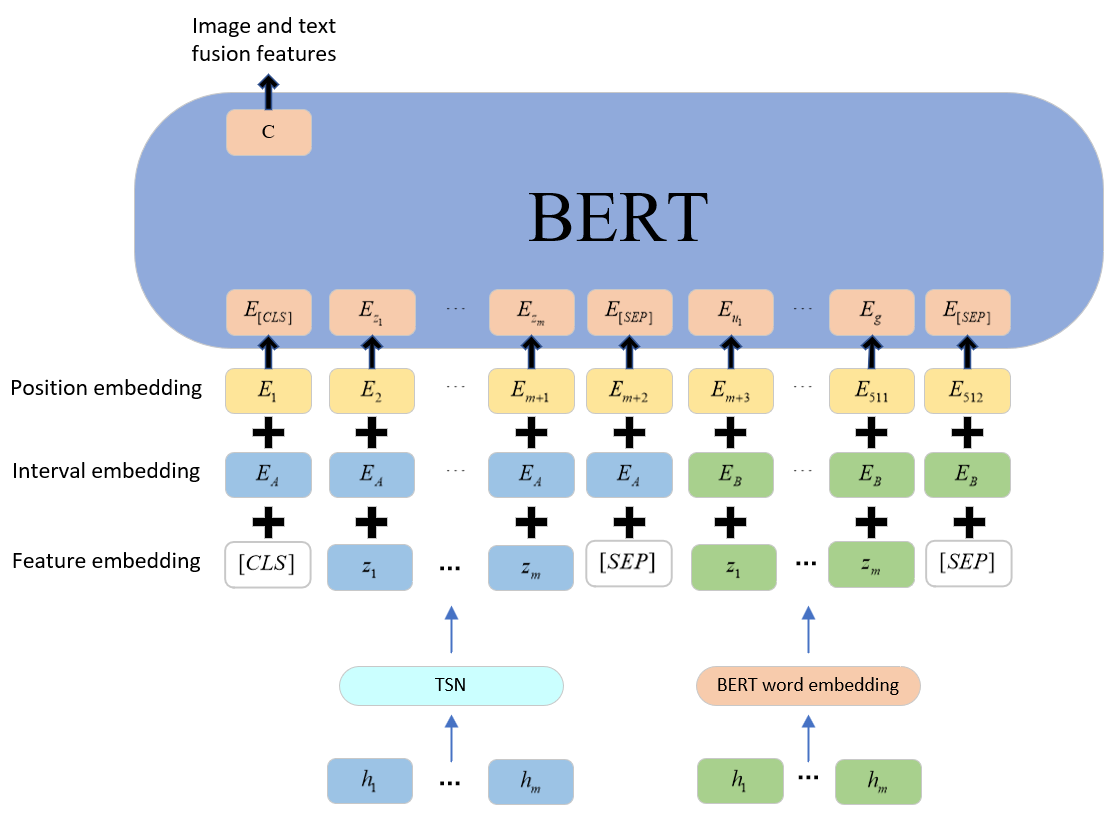}\\
\caption{Input image features and text into the BERT model to get the image and text fusion features}
\end{figure}

The token embeddings, positional embeddings, and segment embeddings are summed and input into the BERT model for fusion. After passing through BERT, the \emph{CLS} token yields a representation vector C, which is a feature vector that fully integrates the image and text features through multiple layers of multi-head self-attention mechanisms, where the text features dominate.

\subsection{Contrastive Learning}

In the original dataset, we have many samples labeled as 0, 1, or 2 (non-rumor, rumor, or debunked). In order to improve the accuracy of retrieval in our vector database, we need to be able to distinguish these samples well in the vector space. We randomly pair the samples into pairs and use them as training data, where the label is 0 if the labels of the two samples are different and 1 if the labels are the same. In this case, we can use contrastive loss: similar pairs with label 1 are pulled together so that they are close in the vector space, while dissimilar pairs that are closer than the defined margin are pushed apart.

In other words, we try to make samples with the same label cluster together in the vector space as closely as possible, while keeping a distance of at least 0.5 cosine similarity between samples with different labels.

We use cosine distance (i.e. 1 minus cosine similarity) as our basic contrastive loss function, with a margin of 0.5. This means that dissimilar samples should have a cosine distance of at least 0.5 (equivalent to a cosine similarity difference of 0.5).

An improved version of contrastive loss is OnlineContrastiveLoss, which looks for negative pairs whose distance is lower than the highest positive pair, and positive pairs whose distance is higher than the lowest negative pair. In other words, this loss automatically detects the hard cases in a batch and only calculates the loss for those cases.

\subsection{Vector search and result prediction}

We obtain feature vectors by feeding existing knowledge into a pre-trained feature representation model, and then insert them into the vector database.

\emph{Approximate nearest neighbor searching} (ANNS) is currently the mainstream approach for vector search. Its core idea is to perform calculations and searches only in a subset of the original vector space, thereby speeding up the overall search speed.

The HNSW(Hierarchical NSW)\cite{malkov2018efficient} and IVF-PQ algorithms are both Approximate Nearest Neighbor (ANN) algorithms. ANN is an approximate approximation of the k-Nearest Neighbor algorithm. In a single vector retrieval problem, we need to find the top-k nearest vectors in the retrieval vector database (gallery) based on a query vector. This is the problem that the kNN algorithm needs to solve. The most intuitive method of kNN is to calculate the similarity between the query vector and all vectors in the retrieval database one by one. In practical applications, the number of vectors in the retrieval database may grow explosively, and brute force kNN search cannot meet the real-time requirements. The purpose of ANN is to sacrifice accuracy for much faster search speed on condition that the precision is allowed. HNSW and IVF-PQ are the two most commonly used methods, where HNSW is graph-based and IVF-PQ is encoding-based.

\begin{figure}[H]
\centering
\includegraphics[scale=.4]{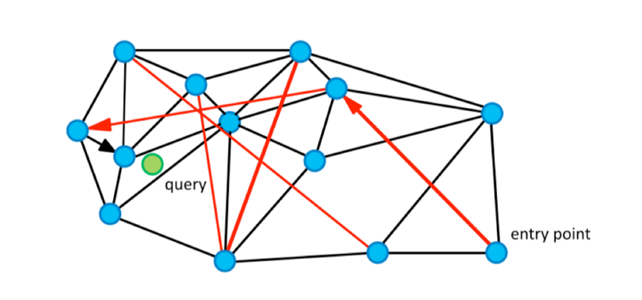}\\
\caption{How to build an NSW graph}\label{NSWGraph}
\end{figure}

HNSW is based on the NSW (Navigable small world models) method. NSW refers to the structure of navigable small world without hierarchy, and building an NSW graph is very simple, as shown in Figure \ref{NSWGraph}. For each new incoming element (the green query node in the figure), we start from a randomly existing point (the entry point node in the figure) and find its set of nearest neighbors (an approximate Delaunay graph) from the structure. As more and more elements are inserted into the structure, the previously used short distance edges become long distance edges, forming a navigable small world.

The search in the NSW graph uses a greedy search approach where the algorithm calculates the distance from the query Q to each vertex in the friend list of the current vertex and selects the vertex with the minimum distance. If the distance between the query and the selected vertex is smaller than the distance between the query and the current element, the algorithm moves to the selected vertex and it becomes the new current vertex. The algorithm stops when it reaches a local minimum: a vertex whose friend list does not contain a vertex closer to the query than the vertex itself.

HNSW (Hierarchical Navigable Small World) is a hierarchical construction of the NSW graph, as shown in the figure. The algorithm uses a greedy strategy to traverse elements from upper layers until a local minimum is reached. Then, the search switches to a lower layer (with shorter connections) and starts searching again from the local minimum in the previous layer. HNSW uses a layered structure to divide edges based on their feature radii, reducing the computational complexity of NSW from polylogarithmic to logarithmic complexity.

\begin{figure}[H]
\centering
\includegraphics[scale=.4]{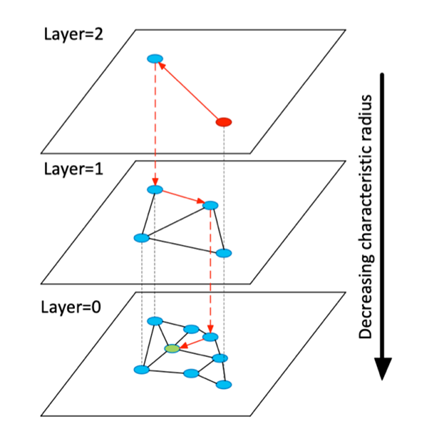}\\
\caption{HNSW algorithm}
\end{figure}

Perform a nearest neighbor search in the vector database and retrieve the top 10 most similar vectors.

For each label (true, false, or debunk video), sum up the cosine similarities of the vectors in the top 10 results that belong to that label. The resulting sum represents the weight of that label. Output the label with the highest weight as the final prediction.

\section{Experiments and results analysis}

This section mainly provides an introduction to the datasets and some basic experimental settings we used, followed by a detailed description of each experiment we conducted. In order to compare the experimental results, we will present the model evaluation metrics on the test set. In addition, we also provide analysis and conclusions for each experiment.

\subsection{The Dataset}

We used the FakeSV short video rumor dataset and a self-annotated short video rumor dataset, the details of which are shown in the table \ref{tab:dataset}.

\begin{table}[]
\begin{tabular}{ll}
\textbf{Type} & \textbf{Quantity} \\
\midrule
Rumors            & 872                   \\
Non-rumors        & 810                   \\
Debunks           & 993                   \\
Total             & 2675                 
\end{tabular}
\caption{Short video rumor dataset}
\label{tab:dataset}
\end{table}

\subsection{Experiment environment}

For this experiment, we used an Nvidia GTX 2080TI 11GB GPU for model training and inference. We stored the video dataset and performed experimental operations on a computer with 32GB RAM, a 500GB solid-state drive, a 4TB mechanical hard drive, and an AMD Ryzen5 5600G processor. The primary programming language environment used for the experiment was Python 3.10.6, and the operating system version was Ubuntu 22.04 LTS.

In the model performance comparison experiment, in addition to the detection method proposed in this paper, we also selected some mature detection methods that perform well in video, audio, and text detection as evaluation baselines, as shown below:
\begin{enumerate}\renewcommand{\labelenumi}{(\theenumi).}
\item Video classification model
\begin{itemize}
\item C3D: The C3D model is a type of 3D convolutional neural network that learns spatio-temporal features of video sequences for feature extraction and classification. The C3D model uses 3D convolution and 3D pooling applied to video frames to capture information from both spatial and temporal dimensions. The C3D model has an efficient, simple, and compact structure, and can achieve good results on multiple video understanding tasks.
\item TSN: The TSN model is a network model used for video classification. It is based on 2D-CNN and extracts global and local information of the video by sparse sampling of video frames. The TSN model contains two branches: the spatial branch and the temporal branch. The spatial branch is used to extract the static features of each frame image, and the temporal branch is used to extract the dynamic features between adjacent frames. The features of the two branches are later fused for classification or detection. The TSN model is a simple and effective video understanding method that can adapt to videos of different lengths and complexity. The TSN model is a video CNN model that uses sparse sampling and dual-branch structure to learn both spatial and temporal features.
\item Slowfast: The Slowfast model is a dual-mode CNN model used for video understanding. It consists of two branches: the Slow branch and the Fast branch. The Slow branch uses fewer frames and larger channel numbers to learn spatial semantic information, and the Fast branch uses more frames and fewer channel numbers to learn motion information. The features of the two branches are fused through lateral connections, and finally classified or detected. The Slowfast model is inspired by the types of retinal nerve cells found in primates, and can effectively utilize different speeds of video information to improve video recognition performance. The Slowfast model is a video CNN model that uses a dual-rate division strategy to learn both spatial and motion features.
\item VideoSwin: The VideoSwin model is a video classification model based on Swin Transformer. It utilizes the multi-scale modeling and efficient local attention properties of Swin Transformer to introduce local induction biases in video Transformer, improving the speed-accuracy trade-off. The VideoSwin model has achieved state-of-the-art performance in video understanding tasks such as action recognition and temporal modeling, surpassing other Transformer models such as ViViT and TimeSformer. The VideoSwin model is a video Transformer model that uses a local attention mechanism to effectively process video sequence data.
\end{itemize}
\item Audio classification models
\begin{itemize}
\item TDNN: Time Delay Neural Network is a neural network used for processing sequential data. It can take multiple frames of feature inputs and extract temporal information through one-dimensional and dilated convolutions. TDNN models were originally used for speech recognition and can build a time-invariant sound model that adapts to speech of different lengths and dialects.
\item ECAPA-TDNN: This model is a network model used for speaker recognition. It is based on TDNN and improves the expression ability of speaker features by introducing techniques such as SE (Squeeze-Excitation) module, multi-layer feature fusion, and channel attention statistical pooling. The ECAPA-TDNN model has achieved excellent performance in multiple speaker verification tasks, surpassing traditional x-vector and ResNet models.
\item Res2Net: This model is a convolutional neural network based on ResNet. It enhances the representation ability and receptive field range of multi-scale features by constructing layered residual connections within the residual block. The Res2Net model divides the input feature map into multiple subsets, each of which undergoes a 3x3 convolution and is added to the output of the previous subset to form a layered connection structure. The Res2Net module can extract time-frequency features of different scales and increase the receptive field range of each network layer, thereby improving the accuracy and robustness of audio classification. Some application scenarios of the Res2Net model for audio classification include speaker recognition, speech emotion recognition, and voiceprint recognition.
\item ResNetSE: This model is a convolutional neural network model that introduces the SE module (Squeeze-and-Excitation module) based on the ResNet model. It enhances the expression ability and selectivity of features by adaptively adjusting the inter-channel weights of the output feature maps of each residual block. The SE module includes two steps: squeeze and excitation. The squeeze step performs global average pooling on each feature map to obtain a channel dimension vector representing global information for each channel. The excitation step performs two fully connected operations on this vector to obtain a channel dimension weight vector representing the importance of each channel. Then, this weight vector is multiplied by the original feature map to obtain the weighted feature map. The ResNetSE model can effectively capture the inter-channel relationship of audio signals, enhance the expression ability and selectivity of audio features, without significantly increasing the computational and parameter overhead.
\item PANNS CNN14: This model is a pre-trained audio neural network used for audio pattern recognition. It has been trained on a large-scale AudioSet dataset. It uses log-mel spectrograms as input features and consists of 14 convolutional layers and 3 fully connected layers.
\end{itemize}
\item Text classification model
\begin{itemize}
\item AttentiveConvNet: This model is a type of convolutional neural network that incorporates attention mechanisms, allowing for dynamic adjustments to the receptive field of the convolutional operation. This results in feature representations that not only contain local context information for each word or phrase, but also global or cross-text information. The AttentiveConvNet model can be applied to natural language understanding tasks such as text classification, textual entailment, answer selection, and others. There are two versions of the AttentiveConvNet model: a light version and an advanced version. The light version uses standard convolutional and pooling layers, weighting feature representations by computing attention weights in each convolutional window. The advanced version builds upon the light version by adding residual connections and multi-head attention mechanisms to enhance feature diversity and expression capability.
\item DPCNN: This model is a text feature extraction and classification model that utilizes a deep pyramid structure and attention mechanisms. It extracts long-range dependencies and abstract features from text by stacking convolutional and pooling layers. The model uses a region embedding layer to convert word embeddings into feature representations that cover multiple words. It employs equally-lengthed convolutional layers to enhance the feature representation of each word or phrase, allowing for a wider range of contextual information to be included. 1/2 pooling layers are used to reduce the length of the text, enlarge the effective receptive field of the convolutional operation, and capture more global information. A fixed number of feature maps are used to reduce computational complexity and maintain semantic consistency. The model uses residual connections to support the training of deep networks and address the issues of vanishing gradients and network degradation.
\item DRNN: This model is a deep recurrent neural network that enhances the model's expressive power and ability to learn long-range dependencies by increasing the number of hidden layers. The main features of the DRNN model include: the repeated repetition of the recurrent unit at each time step, forming a deep recurrent structure. The parameters in each layer of the recurrent unit are shared, but the parameters can differ across layers. The recurrent unit can be any type of recurrent neural unit, such as RNNs, LSTMs, GRUs, and so on.
\item FastText: This model is a text classification and word embedding training tool open-sourced by Facebook AI Research in 2016. Its features include a simple model structure, fast training speed, and satisfactory text classification accuracy. Using all words and corresponding n-gram features as input can capture word order information. The simple average word vector as the text representation can reduce computational complexity. Using hierarchical softmax as the output layer can speed up the training process and is suitable for a large number of categories. Using pre-trained word embeddings as the input layer can improve the effectiveness of text classification.
\item TextCNN: This model is a convolutional neural network used for text classification. The basic idea is to use different sizes of convolutional filters to extract local features from sentences, and then use max-pooling and fully connected layers to obtain the final classification result. It can effectively extract local features from sentences, capturing word order and semantic information. Pretrained word embeddings can be used to enhance the model's generalization ability.
\item TextRNN: The model is a type of recurrent neural network (RNN) used for text classification. Its basic idea is to use RNN to encode the word vector of each word in a sentence into a fixed-length vector, and then input this vector into a fully connected layer to obtain the classification result. It can effectively handle sequence structures and take into account the contextual information of the sentence.
\item TextRCNN: The model is a type of text classification model that combines recurrent neural networks (RNNs) and convolutional neural networks (CNNs). Its basic idea is to use bidirectional RNNs to obtain the contextual information of the sentence, and then concatenate the word vectors and contextual vectors to extract features through a convolutional layer and a max-pooling layer. Finally, a fully connected layer is used to obtain the classification result. It can effectively utilize bidirectional RNNs to capture the contextual information of the sentence, considering word order and semantic information. The max-pooling layer can automatically determine which features are more important in the text classification process, reducing noise and redundant information. Pre-trained word vectors can be used to enhance the model's generalization ability.
\item Transformer: The model is a type of deep learning model based on the attention mechanism, mainly used in natural language processing fields such as machine translation and text summarization. The core idea of the Transformer model is "attention is all you need". It abandons the traditional RNN or CNN structure and uses self-attention mechanism to capture dependencies in the sequence. The Transformer model consists of an encoder and a decoder, each of which contains 6 sub-layers. The encoder is responsible for encoding the input sequence (such as the source language sentence) into a high-dimensional vector representation, and the decoder is responsible for generating the next word based on the encoder's output and the target sequence (such as the target language sentence) that has been generated.
\end{itemize}
We used accuracy, precision, recall, F1 score, ROC curve, and AUC score to evaluate the performance of each neural network model. These evaluation metrics have been widely used in the research community and have become important criteria for evaluating model performance.

For binary classification problems, examples can be classified into four categories based on the combination of their true labels and the predicted labels by the classifier: True Positive (TP), False Positive (FP), True Negative (TN), and False Negative (FN). Let TP, FP, TN, and FN represent the corresponding number of examples. Clearly, TP+FP+TN+FN equals the total number of examples.
\begin{itemize}
\item True Positive (TP), positive examples that are correctly predicted by the model
\item True Negative (TN), negative examples that are correctly predicted by the model
\item False Positive (FP), negative examples that are incorrectly predicted as positive by the model
\item False Negative (FN), positive examples that are incorrectly predicted as negative by the model
\end{itemize}
\end{enumerate}
Accuracy is the most basic evaluation metric in classification problems. It is defined as the percentage of correctly predicted results out of the total number of samples, and its formula is as follows:
$$
\mathrm{Accuracy}=\frac{\mathrm{TP}+\mathrm{TN}}{\mathrm{TP}+\mathrm{TN}+\mathrm{FP}+\mathrm{FN}}
$$
Precision is defined in terms of predicted results. It represents the probability that a sample predicted as positive is actually positive. In other words, it measures how confident we are in predicting positive samples correctly among all predicted positive samples. The formula for precision is as follows:
$$
\mathrm{Precision}=\frac{\mathrm{TP}}{\mathrm{TP}+\mathrm{FP}}
$$
Recall, which is based on the original samples, measures the probability of correctly predicting positive samples among all actual positive samples. Its formula is as follows:
$$
\mathrm{Recall}=\frac{\mathrm{TP}}{\mathrm{TP}+\mathrm{FN}}
$$
The comprehensive evaluation metric F1 score takes into account both Precision and Recall. Sometimes, these two metrics are in a trade-off relationship, meaning that if Precision is high, Recall may decrease, and vice versa. In some scenarios, it is necessary to balance both Precision and Recall. The most commonly used method for this is the F1-Measure, also known as F1-Score, which is the harmonic mean of Precision and Recall. The higher the F1-Score, the better the model's performance.
$$
F_1=2\times \frac{\mathrm{Precision}\times \mathrm{Recall}}{\mathrm{Precision}+\mathrm{Recall}}
$$
As for the triple classification problem of this project, we have to select each of the three labels in turn as positive samples, and then take the other two labels as negative samples to calculate the current TP, TN, FP, and FN, and calculate the accuracy, precision, recall, and overall evaluation index F1 scores respectively based on the previous formula, and then take the average of all the samples, so as to get the final accuracy, precision recall, and F1 value. Specifically, for the triple classification confusion matrix $F_{i,j}$, $F_{i,j}$ is the number of samples whose actual labels are $i$ that are predicted to be $j$, and $N$ is the total number of samples.
$$
\mathrm{Accuracy}=\frac{F_{0,0}+F_{1,1}+F_{2,2}}{N} 
$$
$$
\mathrm{Precision}=(\frac{F_{0,0}}{F_{0,0}+F_{1,0}+F_{2,0}}+\frac{F_{1,1}}{F_{0,1}+F_{1,1}+F_{2,1}}+\frac{F_{2,2}}{F_{0,2}+F_{1,2}+F_{2,2}})\times \frac{1}{3}
$$
$$
\mathrm{Recall}=(\frac{F_{0,0}}{F_{0,0}+F_{0,1}+F_{0,2}}+\frac{F_{1,1}}{F_{1,0}+F_{1,1}+F_{1,2}}+\frac{F_{2,2}}{F_{2,0}+F_{2,1}+F_{2,2}})\times \frac{1}{3}
$$
$$
F_1=2\times \frac{\mathrm{Precision}\times \mathrm{Recall}}{\mathrm{Precision}+\mathrm{Recall}}
$$

ROC(Receiver Operating Characteristic)curve, also known as receiver operating characteristic curve, was first used in the radar signal detection field to distinguish between signal and noise. The horizontal axis of the curve represents FPR (false positive rate), and the vertical axis represents TPR (true positive rate). The diagonal line corresponds to the "random guessing" model, while (0,1) corresponds to the "ideal model" of placing all positive examples before all negative examples. The definitions of FPR and TPR are as follows:

False Positive Rate (FPR), also known as specificity:
$$
\mathrm{FPR}=\frac{\mathrm{FP}}{\mathrm{FP}+\mathrm{TN}}
$$
True Positive Rate (TPR), also known as sensitivity:
$$
\mathrm{TPR}=\frac{\mathrm{TP}}{\mathrm{TP}+\mathrm{FN}}
$$
If the ROC curve of model A is completely enclosed by the ROC curve of model B, then B's performance is better than A's. If the curves of A and B intersect, the model with a larger area under the curve (AUC) has better performance. AUC, also known as the area under the curve, is the size of the area under the ROC curve.

A larger area under the ROC curve indicates better model performance, and AUC is thus an evaluation metric derived from this. Usually, AUC values range from 0.5 to 1.0, with larger values indicating better performance. If a model is perfect, its AUC is equal to 1, indicating that all positive samples are ranked before negative ones. If a model is a simple binary random guess model, its AUC is 0.5. If one model is better than another, its ROC curve will have a relatively larger area under the curve, and the corresponding AUC value will also be larger.

For the ROC curve and AUC area, we define rumor videos as positive samples, and non-rumor and debunking videos as negative samples, thus focusing on the detection performance of rumor videos.

\subsection{Model metrics comparison experiment}

\begin{figure}[H]
\centering
\includegraphics[scale=1]{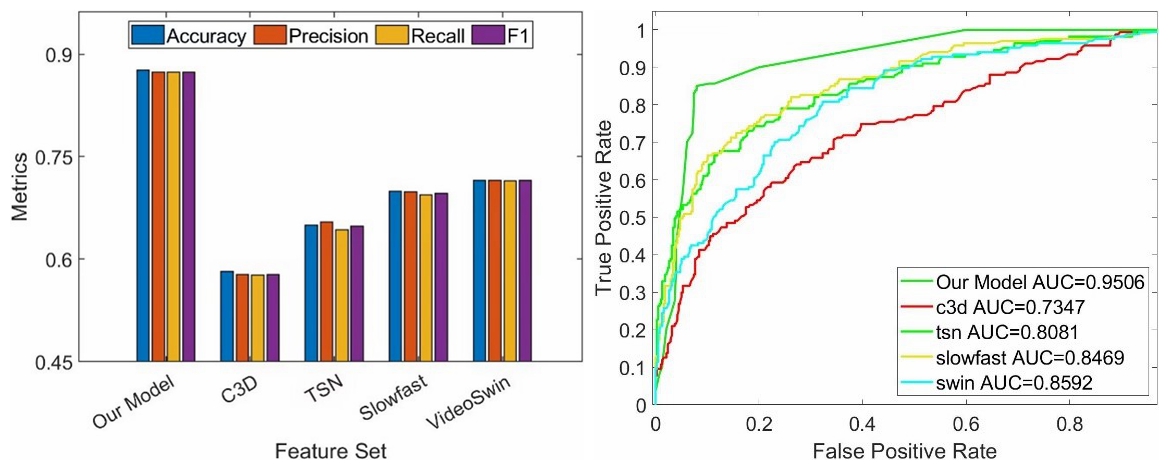}\\
\caption{Comparison results of each index and ROC curve among the video models}\label{videoRes}
\end{figure}

\begin{figure}[H]
\centering
\includegraphics[scale=1]{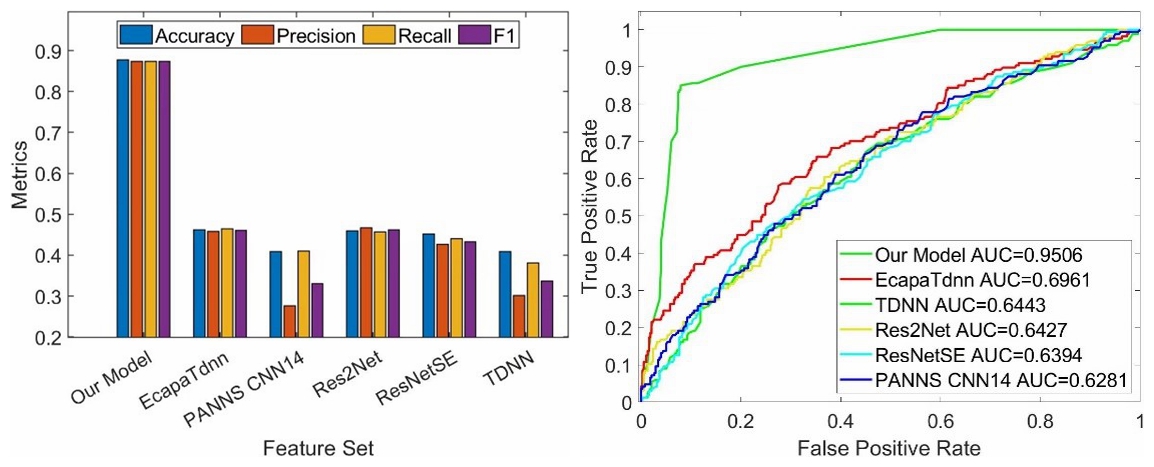}\\
\caption{Comparison results of each index and ROC curve among the audio models}\label{audioRes}
\end{figure}

\begin{figure}[H]
\centering
\includegraphics[scale=1]{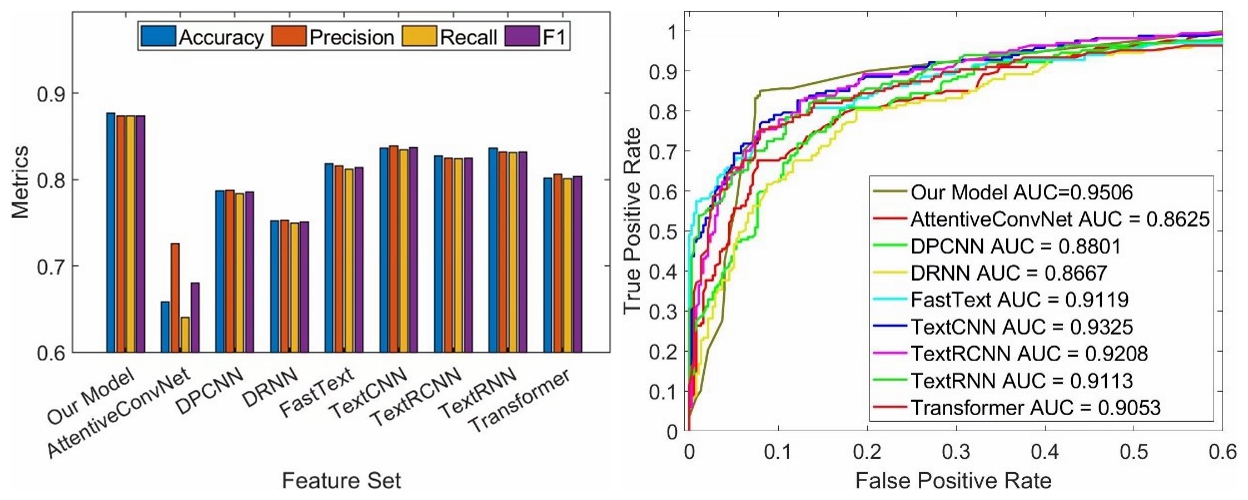}\\
\caption{Comparison results of each index and ROC curve among the text models}\label{textRes}
\end{figure}

The experimental results obtained through training are shown in Table \ref{tab:modelresults} and Figure \ref{videoRes}-\ref{textRes}.

\begin{table}[]
\begin{tabular}{lllllll}
\toprule
Model               & Using Modal             & Accuracy     & Precision     & Recall     & F1-Score    & AUC   \\
\midrule
Our Model        & Graphic+Fusion text+Knowledge & 0.87706 & 0.87374 & 0.87393 & 0.87383 & 0.950670 \\
C3D              & Graphic               & 0.58165 & 0.57746 & 0.57667 & 0.57706 & 0.736083 \\
TSN              & Graphic               & 0.64954 & 0.65418 & 0.64275 & 0.64842 & 0.808954 \\
Slowfast         & Graphic               & 0.69908 & 0.69855 & 0.69416 & 0.69635 & 0.845943 \\
VideoSwin        & Graphic               & 0.7156  & 0.71557 & 0.71471 & 0.71514 & 0.856351 \\
TDNN             & Audio               & 0.40917 & 0.30118 & 0.38096 & 0.33640 & 0.644340 \\
ECAPA-TDNN       & Audio               & 0.46239 & 0.45739 & 0.46438 & 0.46086 & 0.696101 \\
Res2Net          & Audio               & 0.45872 & 0.46723 & 0.45723 & 0.46218 & 0.642735 \\
ResNetSE         & Audio               & 0.45138 & 0.42587 & 0.44082 & 0.43322 & 0.639476 \\
PANNS\_CNN14     & Audio               & 0.40917 & 0.27653 & 0.40966 & 0.33018 & 0.628191 \\
AttentiveConvNet & Fusion text from graphic and audio         & 0.65872 & 0.72618 & 0.64034 & 0.68056 & 0.889364 \\
DPCNN            & Fusion text from graphic and audio         & 0.78716 & 0.7874  & 0.78367 & 0.78553 & 0.884057 \\
DRNN             & Fusion text from graphic and audio         & 0.75229 & 0.75298 & 0.74946 & 0.75121 & 0.869848 \\
FastText         & Fusion text from graphic and audio         & 0.81835 & 0.81612 & 0.81186 & 0.81398 & 0.913997 \\
TextCNN          & Fusion text from graphic and audio         & 0.8367  & 0.83908 & 0.83485 & 0.83696 & 0.928049 \\
TextRCNN         & Fusion text from graphic and audio         & 0.82752 & 0.82526 & 0.82465 & 0.82495 & 0.923154 \\
TextRNN          & Fusion text from graphic and audio         & 0.8367  & 0.83207 & 0.83159 & 0.83183 & 0.918496 \\
Transformer      & Fusion text from graphic and audio         & 0.80183 & 0.80662 & 0.80132 & 0.80396 & 0.907914
\end{tabular}
\caption{Experimental results obtained from training}
\label{tab:modelresults}
\end{table}

The experimental results show that compared with other models, the model proposed in this paper performs better in all three classification indicators when using image+audio fusion, text+external knowledge integration. This indicates that the introduction of multimodality and external knowledge significantly improves accuracy.

\subsection{Model feature representation experiment}

As our final result classification will use vector database retrieval, we hope that the vector similarity between samples of different types is as small as possible, and the vector similarity between samples of the same type is as large as possible. The model's feature representation needs to effectively distinguish between samples of different types.

We obtained vector features of the hidden layer output of all the above models. We use t-SNE dimensionality reduction representation algorithm, which is a nonlinear dimensionality reduction technique that reduces high-dimensional data to two dimensions while preserving local features and distance structures of the data. The basic idea is to transform the distance relationship between high-dimensional data and low-dimensional data into probability distributions and minimize the KL divergence between the two distributions, thereby obtaining a representation of the low-dimensional data.

\begin{figure}[H]
\centering
\includegraphics[scale=.6]{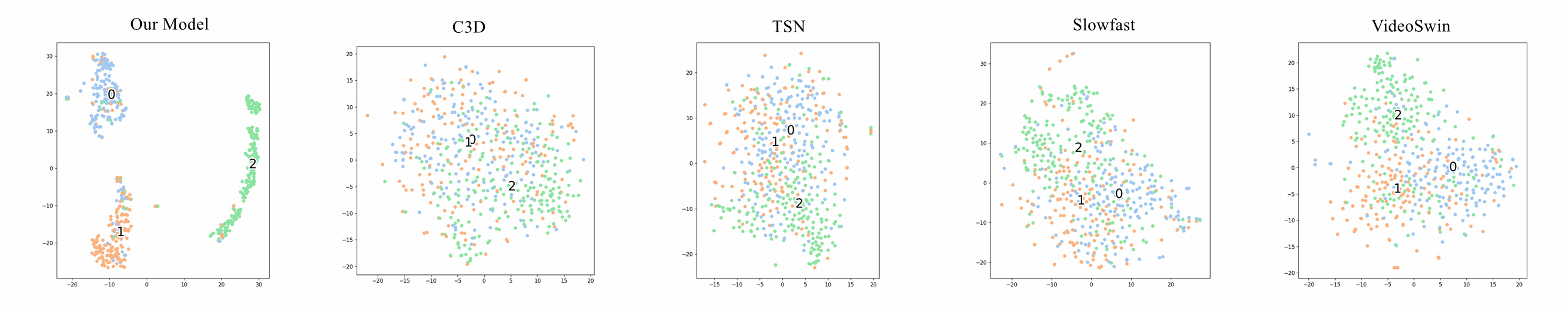}\\
\caption{t-SNE visualization of comparison among resulting vector features of video models}
\end{figure}
\begin{figure}[H]
\centering
\includegraphics[scale=.4]{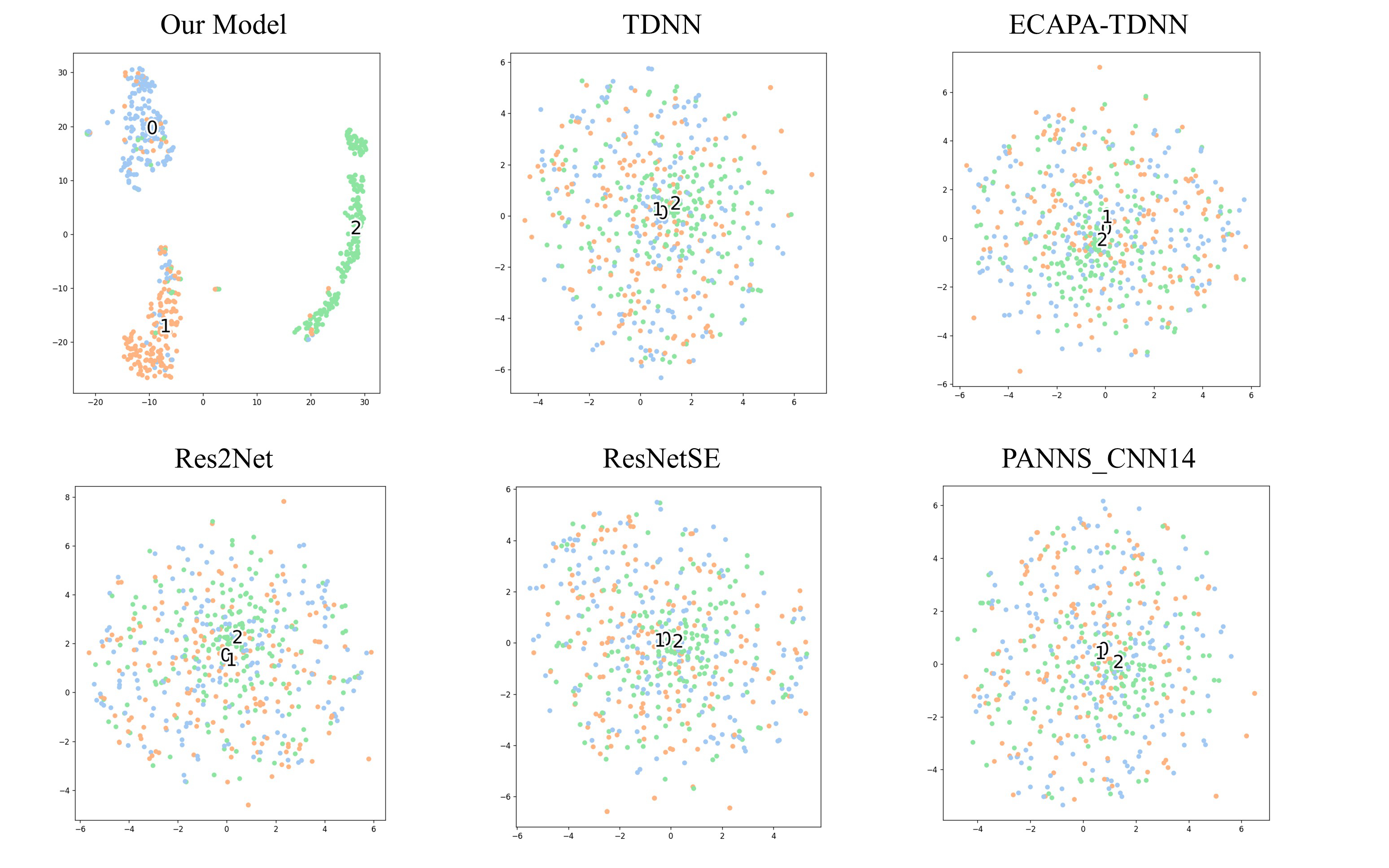}\\
\caption{t-SNE visualization of comparison among resulting vector features of audio models}
\end{figure}
\begin{figure}[H]
\centering
\includegraphics[scale=.1]{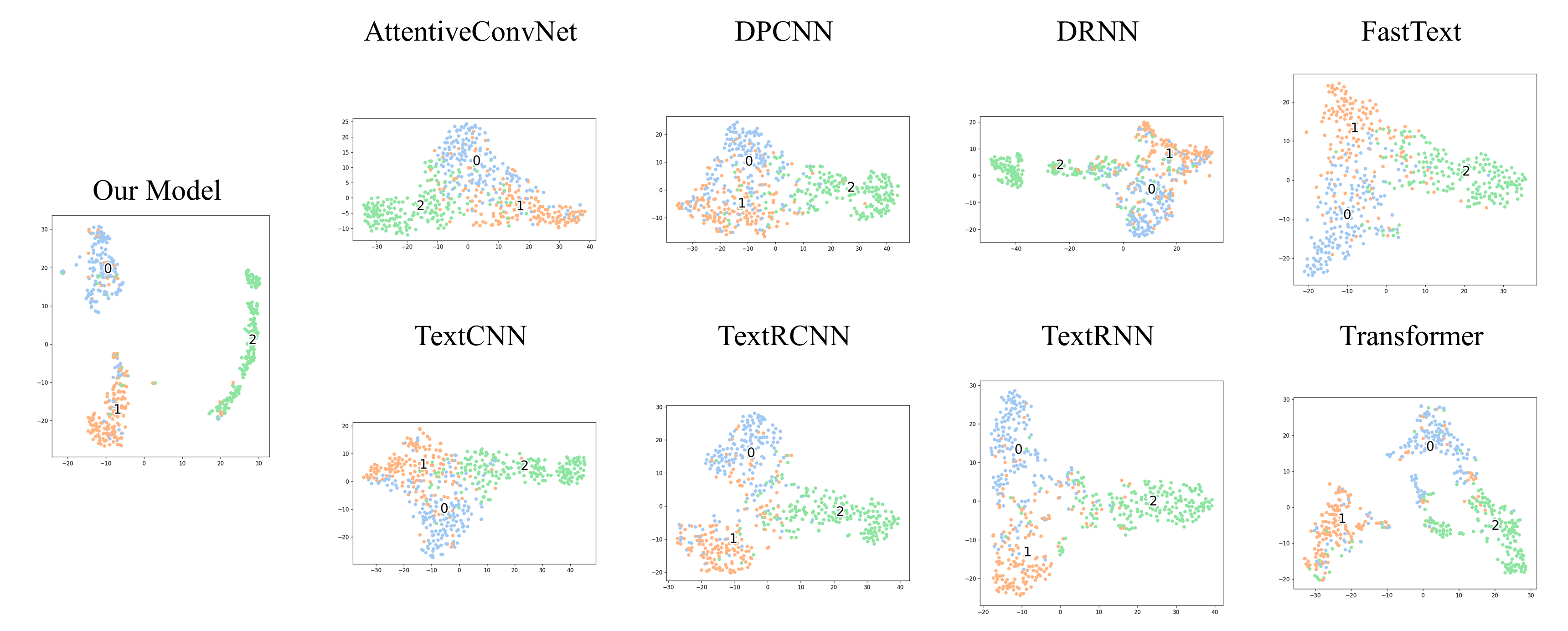}\\
\caption{t-SNE visualization of comparison among resulting vector features of text models}
\end{figure}

From the dimensionality-reduced vector representation features, it can be seen that our model has good discriminative power, with clear clustering areas for each classification, and relatively few misclassified samples. The accuracy of the model's nearest neighbor retrieval is high.

\subsection{Feature ablation experiment}

Feature ablation experiment is a method of evaluating the impact of different features on model performance. It observes the changes in model performance by gradually removing or replacing certain features. 

To explore the importance of each feature in the model, we conducted a feature ablation experiment, and the results are shown in the table \ref{tab:ablationresults}. The table lists different feature combinations and their corresponding performance metrics (such as accuracy, F1 score, etc.). We started with the complete set of features and then gradually removed some features to compare the changes in model performance. We found that the image-text modality feature has the most significant contribution to the model's performance, followed by the RGB image modality feature, the audio-text modality feature has a moderate contribution to the model's performance, while the optical flow modality feature has a lower contribution to the model's performance. The analysis of the movement in the video for rumor analysis is not very clear, and the image text and audio text are complementary. Many videos may have complete textual information in the image, but not all videos may have voiceovers, making it difficult to obtain sufficient textual information from the audio.

\begin{table}[]
\begin{tabular}{llllllll}
\toprule
Image RGB & Image Optical Flow & Image Text & Audio Text & Accuracy     & Precision     & Recall     & F1 Value    \\
\midrule
\checkmark     & \checkmark    & \checkmark    & \checkmark    & 0.87706 & 0.87374 & 0.87393 & 0.87383 \\
\checkmark     &      & \checkmark    & \checkmark    & 0.87339 & 0.87229 & 0.86792 & 0.87010 \\
      & \checkmark    & \checkmark    & \checkmark    & 0.84587 & 0.84176 & 0.84003 & 0.84089 \\
      &      & \checkmark    & \checkmark    & 0.84403 & 0.84120 & 0.83743 & 0.83932 \\
\checkmark     & \checkmark    & \checkmark    &      & 0.86055 & 0.85649 & 0.85656 & 0.85652 \\
\checkmark     & \checkmark    &      & \checkmark    & 0.78899 & 0.78532 & 0.77989 & 0.78260 \\
\checkmark     & \checkmark    &      &      & 0.67156 & 0.67580 & 0.66261 & 0.66914
\end{tabular}
\caption{Characteristic ablation experiment results}
\label{tab:ablationresults}
\end{table}

\subsection{Image and text weighting experiment}

Summarizing video information using key frames is crucial for extracting image features, and the number of image tokens ($m$) is related to the redundancy and coverage of the key frames. If m is too high, although it ensures a large coverage of information, it may cause too much redundancy between frames and the loss of important text features. If m is too small, the key frames cannot summarize the video information, resulting in the loss of important information. Therefore, the optimal number of key frames should be selected to allow the model to fully integrate image and text information, balance the fusion ratio, and achieve the best classification performance.

\begin{figure}[H]
\centering
\includegraphics[scale=0.3]{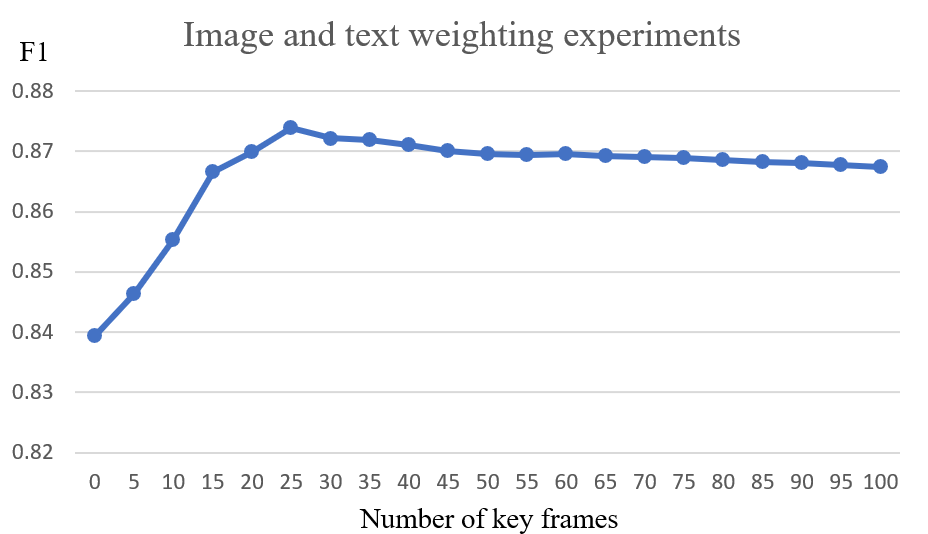}\\
\caption{results of image and text weighting experiments}\label{weightingExp}
\end{figure}

As mentioned earlier, we select $m$ from the range of $0\le m \le 100$ and compare the performance of different $m$ values on the dataset. The curve of the Accuracy metric with the change of $m$ value is shown in Figure \ref{weightingExp}. Overall, the Accuracy metric shows a trend of rapid increase followed by a slow decline. As $m$ increases, the content of image features increases, and Accuracy increases rapidly. After reaching the peak at $m=30$, the coverage of information in key frames is already high enough to summarize video information. However, with the increase of redundancy and the decrease of text information, Accuracy decreases. On the dataset, the model achieves the best Accuracy when $m = 25$; therefore, the model performs best when $m = 25$, i.e., the video accounts for $25$ tokens in the Bert input.

\section{Interpretability (Video Retrieval) Experiment}
We aim to provide the system with relevant reasoning when determining whether a piece of information is a rumor. Therefore, when given a video as input, we intend to locate authentic or debunking videos that are related to the topic of the given video. We have modified the training objective to maximize the similarity between video vectors of the same event and minimize the similarity between video vectors of different events. This allows us to achieve the goal of video retrieval and finding relevant evidence.

\begin{figure}[H]
\centering
\includegraphics[scale=0.3]{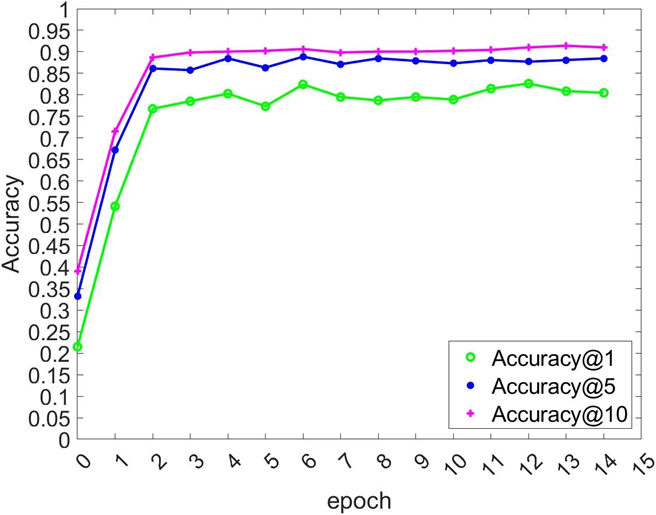}\\
\caption{The training process of the video retrieval model}\label{retrieval}
\end{figure}

As illustrated in Figure \ref{retrieval}, it can be observed that the training results are quite satisfactory, with an accuracy of over $90\%$ in the top $10$ rankings.

\section{Conclusion and future outlook}
In this article, we propose a relatively complete framework for detecting and classifying short video rumors based on multi-modal feature fusion: 1) Dataset establishment: we build a rumor short video dataset with multiple features; 2) Multi-modal rumor detection model: we first use the TSN (Temporal Segment Networks) video encoding model to extract video visual features; then we use the fusion of optical character recognition (OCR) and automatic speech recognition (ASR) technologies to extract text features; next, we use the BERT model to fuse text and video visual features; finally, we use contrastive learning to distinguish between different structures: we crawl external knowledge and introduce it into the vector database for the final structure classification output.

Throughout our research process, we have always been oriented towards practical needs. We hope to continue optimizing our research results and improve the performance of the model in detecting short video rumors. Our goal is to apply our knowledge and achievements to various practical scenarios, such as short video rumor identification and social public opinion control.

\printbibliography

\end{document}